\documentclass[conference]{IEEEtran}

\usepackage{float}
\usepackage{url}
\usepackage{tabularx}
\usepackage{array} 
\usepackage{makecell}
\usepackage{multirow}
\usepackage{booktabs}
\usepackage[colorlinks,urlcolor=blue,linkcolor=blue,citecolor=blue]{hyperref}
\usepackage[dvipsnames,table]{xcolor}
\definecolor{LightCyan}{rgb}{0.88,1,1}
\usepackage[justification=centering]{caption}
\usepackage{balance}
\usepackage{graphicx} 
\usepackage{adjustbox} 
\usepackage{comment}
\usepackage{cite}
\usepackage{amsmath,amssymb,amsfonts}
\usepackage{algorithmic} 
\usepackage{textcomp}

\DeclareRobustCommand{\IEEEauthorrefmark}[1]{\smash{\textsuperscript{\footnotesize #1}}}

\newif\ifcomments
\commentstrue

\begin{document}

\title{LLMZ+: Contextual Prompt Whitelist Principles for Agentic LLMs}

\author{
    \IEEEauthorblockN{
    Tom Pawelek\IEEEauthorrefmark{1},
    Raj Patel\IEEEauthorrefmark{2},
    Charlotte Crowell\IEEEauthorrefmark{2},
    Noorbakhsh Amiri Golilarz\IEEEauthorrefmark{2},\\
    Sudip Mittal\IEEEauthorrefmark{2},
    Shahram Rahimi\IEEEauthorrefmark{2}, and
    Andy Perkins\IEEEauthorrefmark{1}
    }\\
    
    \IEEEauthorblockA{\IEEEauthorrefmark{1}Mississippi State University, Mississippi State, MS, USA}
    \IEEEauthorblockA{\IEEEauthorrefmark{2}The University of Alabama, Tuscaloosa, AL, USA}\\
    
}


\maketitle

\begin{abstract}
    Compared to traditional models, agentic AI represents a highly valuable target for potential attackers as they possess privileged access to data sources and API tools, which are traditionally not incorporated into classical agents. Unlike a typical software application residing in a Demilitarized Zone (DMZ), agentic LLMs consciously rely on nondeterministic behavior of the AI (only defining a final goal, leaving the path selection to LLM). This characteristic introduces substantial security risk to both operational security and information security. Most common existing defense mechanism rely on detection of malicious intent and preventing it from reaching the LLM agent, thus protecting against jailbreak attacks such as prompt injection.
    In this paper, we present an alternative approach, LLMZ+, which moves beyond traditional detection-based approaches by implementing prompt whitelisting. Through this method, only contextually appropriate and safe messages are permitted to interact with the agentic LLM. By leveraging the specificity of context, LLMZ+ guarantees that all exchanges between external users and the LLM conform to predefined use cases and operational boundaries. Our approach streamlines the security framework, enhances its long-term resilience, and reduces the resources required for sustaining LLM information security. Our empirical evaluation demonstrates that LLMZ+ provides strong resilience against the most common jailbreak prompts. At the same time, legitimate business communications are not disrupted, and authorized traffic flows seamlessly between users and the agentic LLM. We measure the effectiveness of approach using false positive and false negative rates, both of which can be reduced to 0 in our experimental setting.

\end{abstract}

\begin{IEEEkeywords}
security, LLM, Prompt injection, Prompt jailbreak, Agentic AI, LLMZ+, Prompt whitelisting, Contextual whitelisting 
\end{IEEEkeywords}

\section{Introduction}

  Throughout commercial and academic applications, there are numerous products, studies and libraries designed to prevent malicious prompts from interacting with LLMs in production environments. All of these approaches operate on the principle of identifying malicious messages and blocking their progression within the system. The inherent structure is similar to that of traditional anti-malware products, relying on a predefined set of signatures and heuristics. Since all detection is driven by these definition databases, they must be updated regularly. When a new attack technique emerges, the definitions must be revised to ensure that the system can recognize the new threat.
  
  These products, therefore, introduce additional capital expenditures (CapEx) \cite{nist2020IR} and operating expenses (OpEx) \cite{nist2020IR}, as their maintenance requires specialized resources. They also pose a hidden risk of ``failing silently”, whereby outdated or delayed definition updates can result in the system providing a false sense of security while remaining vulnerable to newly developed attacks. Our objective was to design a framework that is more accommodating for administrators and is based on a security principle that leverages the contextual information of the subject being served by the agentic LLM.

  In developing our solution, we drew inspiration from the most secure practices employed in perimeter firewall design. Rather than maintaining an exhaustive database of all global threats (sources, payloads, etc.), the recommended strategy is to allow only those types of traffic that are explicitly recognized as safe and legitimate, while blocking all other network packets \cite{nist2009}. As a result, a typical configuration specifies the scenarios in which access is permitted (such as traffic originating from domestic IP addresses, residential internet service providers, or specific destination ports) which is then concluded with a catch-all rule: $DENY ALL$.

  This approach relieves network administrators from the burden of maintaining an ever-growing list of foreign addresses, Virtual Private Network (VPN) \cite{nist_vpn} subnets, exploited ports and daemon processes \cite{gfg_daemon_processes}, and and other threat definitions. Instead, the system operates on the principle that only recognized and compliant traffic is permitted. We adopt this foundational principle and apply it to the design of our LLM guard, as described in the following sections.

  In this paper, we present \textit{LLMZ+}: a solution designed to safeguard agentic AI models against prompt injection attacks. Prompt injection is a form of jailbreak attack in which malicious prompts are used to bypass built-in checks and override the control mechanisms of LLMs \cite{nist2025}. Our approach introduces a contextual whitelisting mechanism that is grounded in a comprehensive understanding of analyzed messages, ensuring that only those prompts relevant to the intended use case are permitted. The major contributions of this paper are as follows:
    \begin{itemize}
        \item We introduce \textbf{\textit{LLMZ+}} (LLM + DMZ \cite{nist_dmz}) as a conceptual security boundary for agentic LLMs, drawing inspiration from the Demilitarized Zone (DMZ) architecture in network security.
        \item We highlight the limitations of conventional prompt threat detection and mitigation techniques, which often rely on static heuristics and often struggle to counter adaptive adversaries.
        \item We introduce a framework tailored for business-focused agentic LLMs, designed for both operational reliability and security resilience.
        \item We evaluate LLMZ+ using a benchmark of documented prompt injection attacks, alongside authentic business communications submitted to agentic chatbots.
    \end{itemize}

    The remainder of this paper is structured as follows. Section II reviews the current state-of-the-art and related work. Sections III and IV introduce the LLMZ+ framework and detail the threat model addressed in this study. Sections V through VII present our experimental setup, results, and practical considerations for deploying LLMZ+ in production business environments. Finally, Section VIII offers conclusions and outlines directions for future research.

\section{Background and Related work}

    In this section, we provide the background on previously explored jailbreaking techniques within LLM as well as defenses that have been presented to combat against it. The phenomenon of jailbreaking in LLMs refers to the circumvention of built-in safety mechanisms by crafting adversarial prompts that elicit responses normally restricted by the system’s alignment objectives \cite{liu_2024}. Early work by Wei et al. \cite{wei_2023JailbrokenLLMfirst} introduced Greedy Coordinate Gradient Ascent (GCG), a method using suffix-based mismatch objectives and alternate encodings (e.g., base64) to bypass safety filters. Building on this, Zou et al. \cite{zou2023universaltransferableadversarialattacks} extended the attack by combining multi-prompt, multi-model strategies to create universal jailbreak prompts.
    
    Several studies have analyzed the taxonomy of jailbreak strategies. Liu et al. \cite{liu_2024} categorized jailbreak prompts into three core patterns: pretending, attention shifting, and privilege escalation. Their study found that both ChatGPT-3.5 and GPT-4.0 were vulnerable to these methods, with an 86\% success rate. Gupta et al. \cite{gupta_2023} identified four dominant jailbreak vectors: role-playing (e.g., Do Anything Now (DAN) or Developer Mode), reverse psychology, model escape, and prompt injection. These techniques can be exploited to generate content for malicious purposes such as phishing, social engineering, and malware development. Similarly, Yu et al. \cite{yu_2024dontListenToMe} conducted a qualitative study involving user-generated prompts, showing that even untrained users were capable of producing effective jailbreaks. Their work also introduced a hybrid human-AI prompting framework, though they noted the AI component's difficulty in adapting to semantic nuance.
    
    Beyond prompt engineering, Carlini et al. \cite{carlini_2023arealignedneuralnetworksadversariallyaligned} demonstrated that brute-force strategies and adversarial multi-modal inputs (e.g., malicious images) could induce harmful outputs, suggesting a broader surface for attack. Complementing these studies, Chen et al. \cite{chen_2024} proposed a suffix-classification model that successfully identified and blocked GCG-style jailbreaks with 96\% accuracy. 
    
    The challenges extend to multilingual and synthetic language models as well. Deng et al. \cite{deng2024multilingualjailbreakchallengeslarge} found that LLMs are more vulnerable in low-resource languages, while Oh et al. \cite{ascii+7_2025} showed that malicious prompts encoded in synthetic ASCII-based formats could also bypass content filters.      
    
    From a system-wide perspective, Yao et al. \cite{yao_2024} conducted a large-scale survey on the intersection of LLMs and security, categorizing existing literature into three thematic areas: the use of LLMs for defensive cybersecurity applications (``good"), the misuse of LLMs for offensive purposes such as cyberattacks (``bad"), and studies addressing inherent vulnerabilities in LLMs along with corresponding defense mechanisms (``ugly"). Separately, Das et al. \cite{das_2025} provided an in-depth analysis of LLM-specific vulnerabilities, including jailbreaking, data poisoning, and personally identifiable information (PII) \cite{nist2015} leakage, and surveyed a broad range of proposed mitigation strategies.
    
    Among emerging defenses, Reinforcement Learning from Human Feedback (RLHF) has been foundational in improving LLM alignment \cite{ouyang2022traininglanguagemodelsfollowRLHF}, with Ganguli et al. \cite{ganguli2022redteaminglanguagemodels} further emphasizing red teaming as a critical evaluation method. Additional efforts like \textit{SmoothLLM} \cite{robey2023smoothllm}, \textit{LLM Guard} by Protect AI \cite{protectai2024llmguard}, and Deng et al.'s \textit{SELF-DEFENCE} framework \cite{deng2024multilingualjailbreakchallengeslarge} employ strategies such as prompt sanitization, multilingual safety data generation, and adversarial fine-tuning. Hila et al. \cite{gonen2024demystifyingpromptslanguagemodels} proposed reducing prompt perplexity through translation by a secondary LLM to improve resilience.
    
    Despite these efforts, a consistent pattern emerges in that most defenses rely on either input and output prompt filtering or RLHF-based alignment techniques. Input filtration typically targets jailbreak attempts, while output filtration aims to prevent the disclosure of Personally Identifiable Information (PII), or other sensitive content. These approaches often depend on static mechanisms, such as maintaining and updating keyword lists or periodically retraining models, which may limit their adaptability and responsiveness. RLHF, while effective, is costly, time-consuming, and requires frequent retraining to remain relevant against evolving attack strategies. 
    
    By contrast, our approach, termed \textbf{LLMZ+}, is highly restrictive in its behavior, continuously context-aware, and capable of dynamically enforcing alignment during inference. It actively monitors whether a prompt deviates from the operational domain, and any deviation triggers an immediate denial of the response. This persistent, context-aware, real-time validation distinguishes our framework from prior defenses that depend on static filters or periodic model updates. In the next section, we elaborate on the core architecture and enforcement logic underpinning the LLMZ+ framework.

\section{Principles of LLMZ+}

\begin{figure}[!ht]
        \centering
        \includegraphics[scale=0.75]{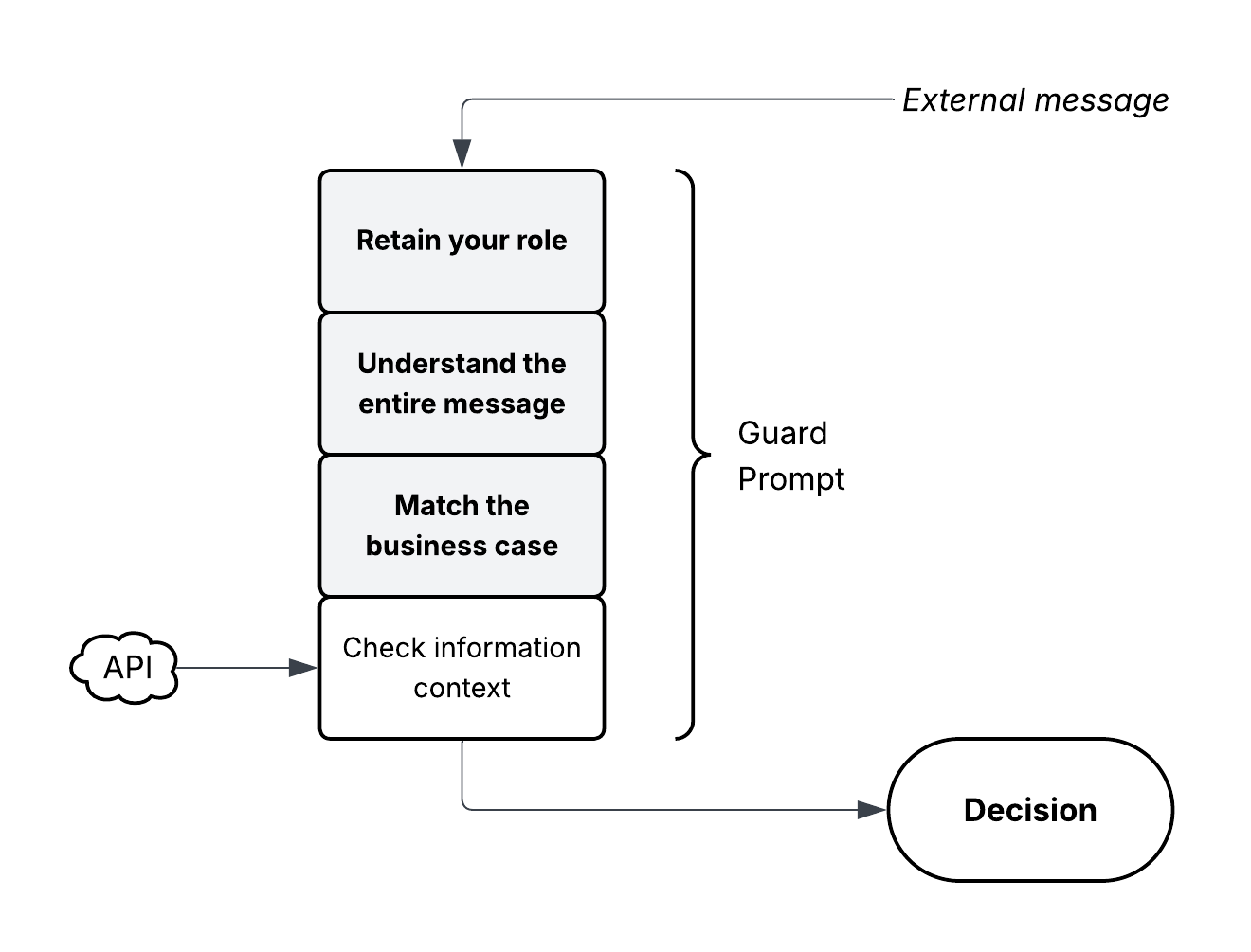}
        \caption{LLMZ+ Guard Prompt Structure}
        \label{fig:sec-prompt}
    \end{figure}

    In our context-based approach, we employ a Guard Prompt, as depicted in Fig. \ref{fig:sec-prompt}, to evaluate all incoming external messages. Instead of searching for prompt exploits, which may include a variety of linguistic and mathematical manipulations, our method ensures that every message is fully understood by the guarding LLM and corresponds to the expected use case of the ongoing conversation. To further protect outbound messages produced by the agentic AI, an information scope layer can be incorporated. This additional protection may be integrated directly within the LLM or implemented through established Data Loss Prevention (DLP) mechanisms \cite{nist2023}. The information scope explicitly defines which categories of information, such as specific types of PII, the model is authorized to disclose. For instance, this restriction may only permit the model to return information that is necessary for a customer to access their account.

    This approach protects the agentic LLM from a wide range of exploitive prompts and mitigates the risk of both known and novel prompt injection attacks. By leveraging the specific deployment context of the agent, the scope of acceptable content exchanged with the agentic core is further narrowed, reducing the potential attack surface.

    The result of the Guard Prompt evaluation may be represented as either a binary decision indicating yes or no, or as a quantified risk score ranging from 0 to 10 for more nuanced applications. Ideally, all malicious messages would be assigned a risk score of 10, while all benign messages would receive a score of 0.

\section{Threat Model}
    In this section, we define the specific threat model targeted by the LLMZ+ framework. We begin by describing typical deployment scenarios for agentic LLMs in practical business contexts. Next, we analyze major attack vectors relevant to these deployments and detail the potential risks they introduce. We then present our proposed security solution, including both its architectural design and deployment strategy. This overview establishes a foundation for understanding the security objectives, effectiveness, and boundaries of the LLMZ+ approach. 
    
    \subsection{Typical Agentic LLM Deployment}

    In our analysis, agentic AI is frequently deployed as a customer-facing LLM (via web chat, phone audio or mobile text). In commercial settings, these agentic models are designed to serve specific use cases, such as:

    \begin{itemize}
        \item Providing customer support,
        \item Facilitating payments,
        \item Assisting with product or service selection.
    \end{itemize}

    It is important to note that our solution does not attempt to secure generic, all-purpose agents that are publicly accessible and context-agnostic, where the nature of each interaction is determined solely by the end-user. Such systems are usually not hosted within a corporate Demilitarized Zone (DMZ) \cite{nist_dmz}, which is an isolated network segment positioned between internal and external environments and commonly used to provide controlled access to on-premise services. As a result, these agents generally lack privileged access to sensitive data sources or APIs. In contrast, agentic LLMs designed for targeted business tasks often require access to confidential, non-public information in order to complete their assigned task.

    \subsection{Attack Vectors}
    Given the privileged access held by these agentic LLMs, attackers may attempt to exploit prompt engineering techniques to ``jailbreak” the model and gain unauthorized control over sensitive information. For an agentic LLM integrated with multiple APIs, manipulating the model to perform arbitrary API calls and reveal the results is analogous to obtaining shell access to a compromised server. The consequences of such breaches are similar to those of traditional network intrusions and typically manifest in two primary forms:

    \begin{itemize}
        \item \textbf{Sensitive data leakage:} The compromised LLM may be used to extract personally identifiable information (PII), financial records, trade secrets, or other confidential data \cite{nist2024}.
        \item \textbf{Unauthorized activity:} Attackers can induce the LLM to execute actions resulting in material harm, such as transferring funds, sending unauthorized communications, or disrupting system operations.
    \end{itemize}

    Although one might assume that the impact is confined to the data and tools directly accessible to the LLM, in practice, these attacks often serve as an initial foothold for further lateral movement within the organization’s network. This type of exploitation can bypass traditional security controls, including firewalls and DLP systems.

    Attackers typically target the LLM through public interfaces, without requiring any privileged access or insider knowledge of the AI deployment. For the purposes of this study, we restrict our focus to prompt-based attacks and do not address other forms of network or software compromise. Accordingly, LLMZ+ is not intended to replace a comprehensive information security (infosec) architecture \cite{nist2017}; instead, it serves as an additional safeguard that protects the LLM from prompt-based adversarial attacks.

    \subsection{Proposed Solution}
\begin{figure}[!ht]
        \centering
        \includegraphics[scale=0.5]{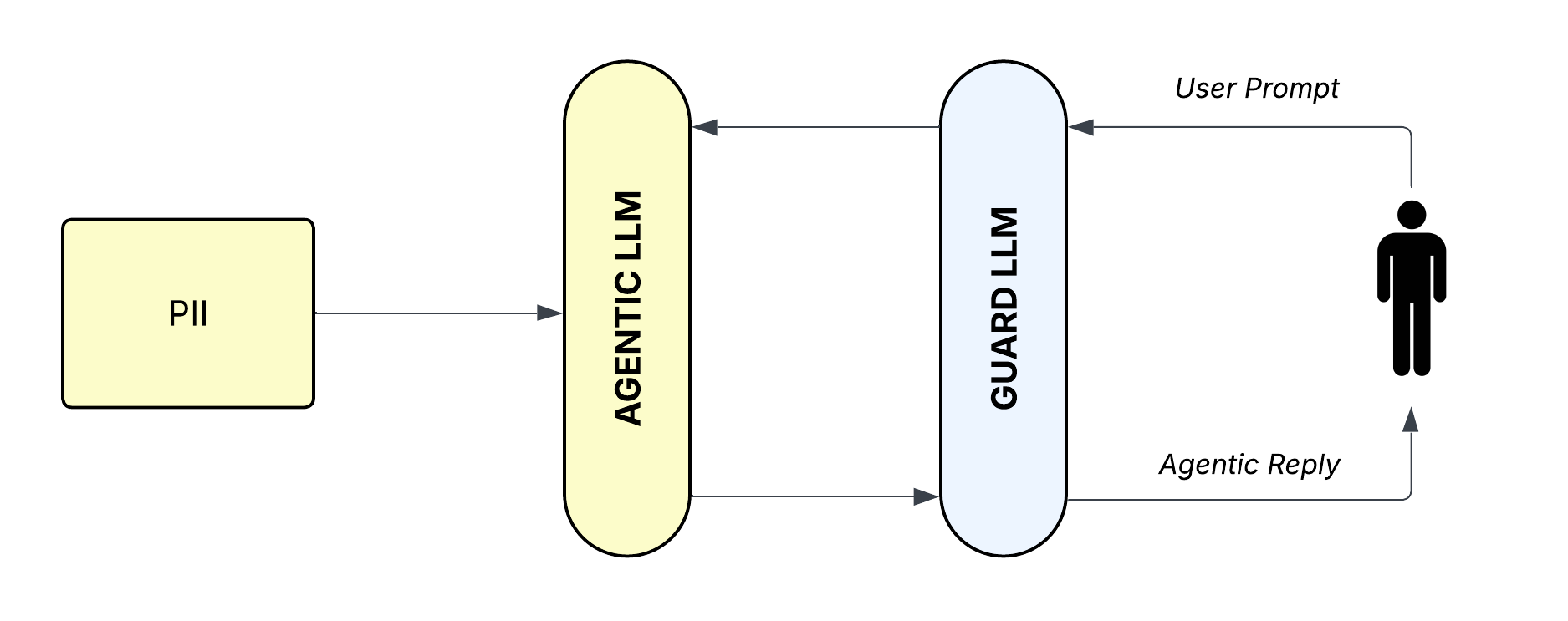}
        \caption{LLMZ+ Data Flow}
        \label{fig:sec-data}
    \end{figure}

    As shown in Fig. \ref{fig:sec-data}, our approach leverages an auxiliary LLM (in blue), to function as a whitelist guard for both ingress prompts sent from external users, as well as egress replies returned by the agentic LLM (in yellow). Following the ``Firewall principle", instead of attempting to enumerate and block every possible malicious input, LLMZ+ evaluates each message against a set of strict criteria.

    The \textbf{Ingress filter} verifies that messages received from external users meet the following requirement, such as:

        \begin{itemize}
            \item The message is fully interpretable by the Guard Prompt.
            \item The message is consistent with a natural customer-service conversation.
            \item The message is relevant to the business case served by the Agentic LLM.
        \end{itemize}

    The \textbf{Egress filter} ensures that outbound messages also remain consistent with the intended business use case. To enhance our safety mechanism, we can incorporate a very simple contextual Retrieval-Augmented Generation (RAG), which informs the guard LLM about the categories of data that the user is permitted to access (e.g. account information, balance, transactions, etc.). A simpler variant uses the information whitelisting implemented directly in the prompt or regex-based filter to block the disclosure of sensitive information like Social Security Numbers (SSNs).

    Messages in either direction that do not satisfy these criteria are blocked, preventing the exploitation of the agentic LLM through prompt-based attacks. It is important to emphasize that this solution is designed specifically to address prompting threats and does not mitigate risks associated with other layers of the AI deployment stack, such as network security vulnerabilities or traditional software exploits.

    \subsection{Deployment Strategy}
    Our implementation is evaluated in the context of a commercial fintech chatbot \cite{fintech_chatbots} deployed within a highly regulated retail market in the US. In this deployment, when a suspicious message is intercepted by LLMZ+, the system notifies a human operator who can then review the communication or directly intervene in the conversation as necessary.

    To assess the effectiveness of our approach compared to traditional LLM threat detection methods, we constructed a controlled testing environment based on an on-premises setup that includes the following components:

    \begin{itemize}
        \item 2x Llama3.1 \cite{ollama_llama3.1} / Llama3.3 \cite{ollama_llama3.3} models (agentic prompts + guard prompts)
        \item OpenWebUI \cite{openwebui}
        \item A static data source and dynamic API accessible to the Agentic LLM
    \end{itemize}
    
    The primary task in this scenario involves performing customer account login and balance confirmation. This represents a very simple use case, albeit it can be implemented in an agentic fashion, using a set of OpenAPI calls depending on how our customers decide to authenticate (account number, SSN or a phone lookup). This scenario is an example where case specificity is used as a foundation of context white-listing. It can be easily generalized to any similar business use.

    We conducted two types of evaluations. First, we sent a representative set of legitimate customer messages through LLMZ+ to measure the rate of \textbf{false positives} rates. Second, we utilized a public repository of ``GPT Super Prompting" \cite{awesome_gpt_super_prompting} techniques, which contains the most recent jailbreak techniques designed to fool LLMs into acting against the constraints defined by their authors. Our goal was to measure how many of those prompts would be blocked by LLMZ+, and how many would pass through (i.e. the \textbf{false negative} rates).  

    Attack attempts against whitelisting-based filters are comparatively rare. key advantage of our approach is the clear separation of user-supplied messages from the rest of the LLM prompt, which makes it significantly more difficult for adversaries to conceal and deliver malicious instructions that might be executed by our agent.

\section{Methodology}
    Our system setup is depicted in Fig. \ref{fig:sec-infa} and described in detail in ``Deployment Strategy" in the previous sub-section. 
    The primary objective of our study is to minimize false rates, ideally reducing them to zero, which would indicate an optimally configured system with full alignment between quantified and binary evaluation scores. The false positive and false negative ratios were experimentally measured as follows:

    \begin{figure}[!ht]
        \centering
        \includegraphics[scale=0.49]{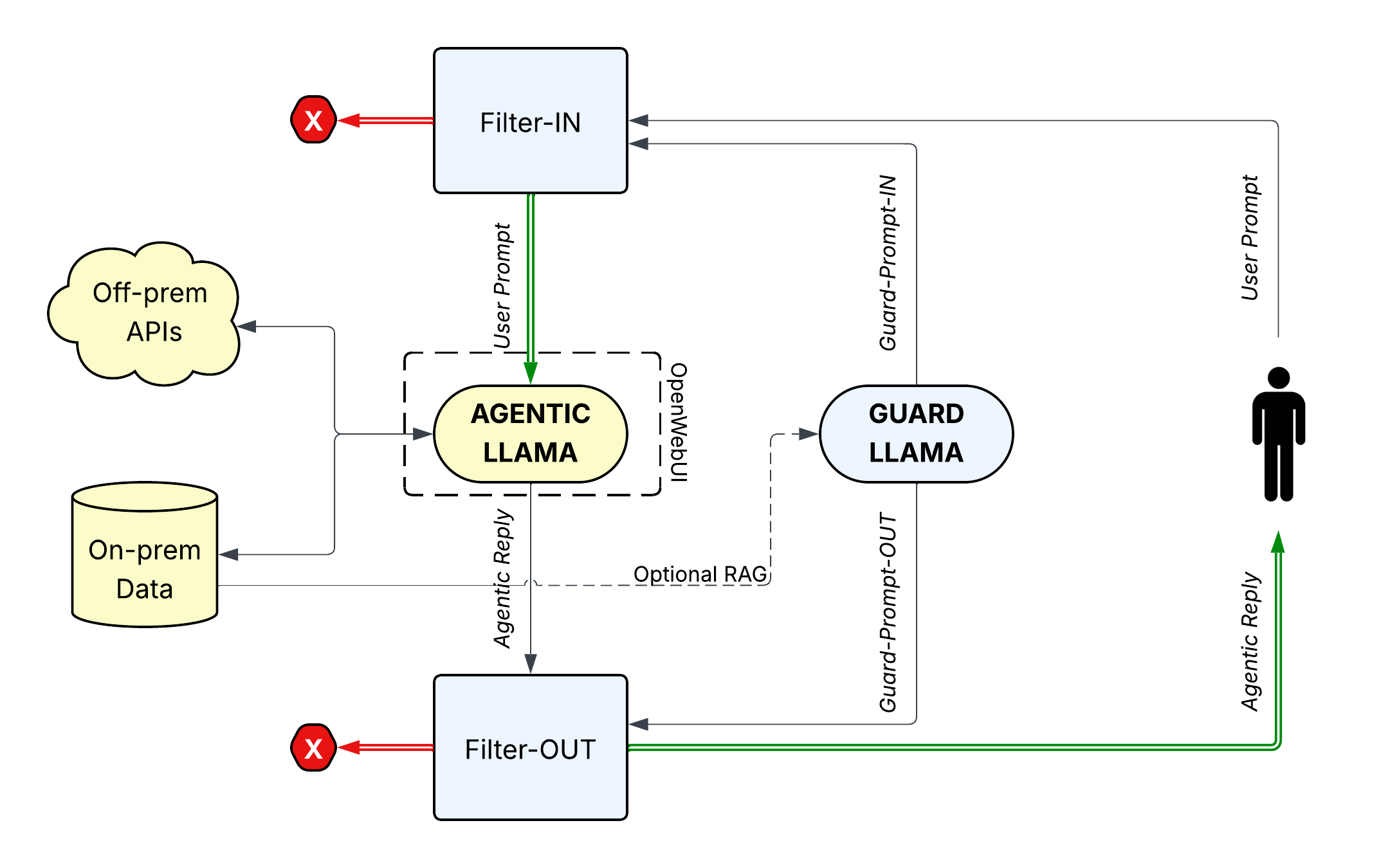}
        \caption{LLM Infrastructure Setup}
        \label{fig:sec-infa}
    \end{figure}

    \subsection{False Negatives}
    False negatives are defined as malicious messages that are erroneously allowed to be processed by the agentic LLM. This rate serves as a key indicator of the system’s effectiveness in threat mitigation. To specifically evaluate the core performance of LLMZ+, we did not incorporate a Retrieval-Augmented Generation (RAG) \cite{nist_rag} in our tests.
    
    A set of scripts was developed to simulate user inputs, utilizing prompts from \cite{awesome_gpt_super_prompting}. These prompts were submitted to a RESTful API endpoint \cite{aws_restful_api}, which served as the backend for a text-based chatbot. The false negative rate was calculated as follows:

    \begin{equation}
    F_{\text{negative}} = 1 - \frac{C_{\text{flagged}}}{C_{\text{total}}}
    \label{eq:negative_fraction}
    \end{equation}

    where:

    \begin{itemize}
        \item F\textsubscript{negative} refers to the false negative rate 
        \item C\textsubscript{flagged} refers to the count of messages flagged by LLMZ+
        \item C\textsubscript{total} refers to the total count of messages posted
    \end{itemize}

    \subsection{False Positives}
    False positives are legitimate messages that do not aim to exploit the agentic LLM, but are incorrectly flagged as malicious by LLMZ+. While not a direct information security threat, a high false positive rate could discourage commercial adoption due to the risk of business disruption and degraded customer service.

    To evaluate the false positive rate, we compiled a collection of authentic end-user messages from agentic bots deployed in business environments similar to our experimental context. These scenarios required customers to authenticate in order to access restricted information, such as PII. In these cases, we expect LLMZ+ to allow all legitimate messages to pass through. The false positive rate was determined as follows:

    \begin{equation}
    F_{\text{positive}} = \frac{C_{\text{flagged}}}{C_{\text{total}}}
    \label{eq:positive_fraction}
    \end{equation}

    where:

    \begin{itemize}
        \item F\textsubscript{positive} refers to the false positive rate,
        \item C\textsubscript{flagged} refers to the count of messages flagged by LLMZ+,
        \item C\textsubscript{total} refers to the total count of messages posted.
    \end{itemize}

    \subsection{Decision Threshold (DT)}
    For the purpose of our experiment, we define Decision Threshold (DT) as a cut-off value. Any messages assigned a score equal to or greater than DT are stopped by LLMZ+ and are not forwarded to the agentic AI worker.

\section{Results and Observations}
\begin{figure}
        \centering
        \includegraphics[scale=0.48]{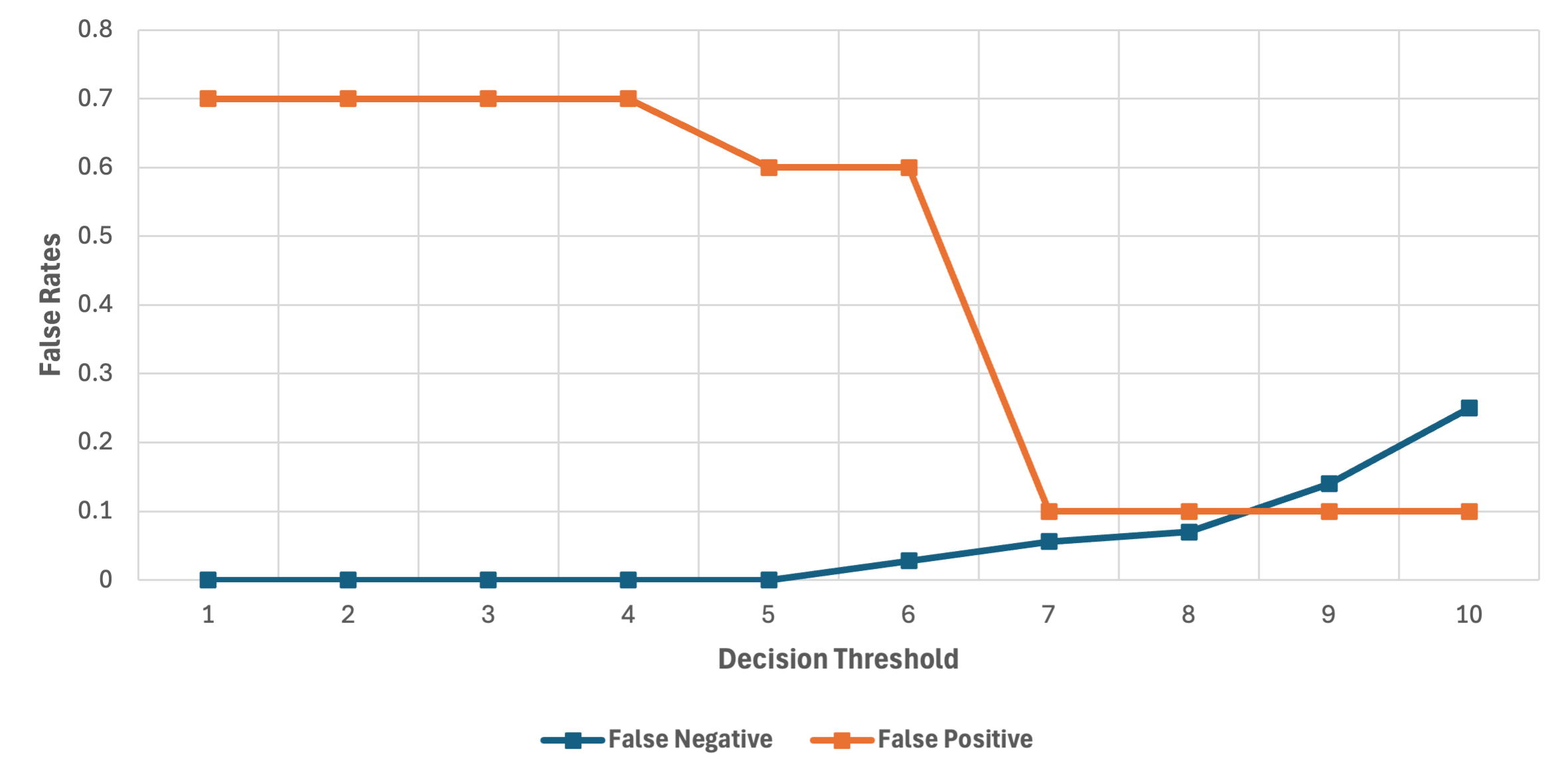}
        \caption{False rates, $Llama3.1_{8B}$, $C_{tot}$ = 71}
        \label{fig:fn-Llama31-8b}
    \end{figure}

    To facilitate risk-based decision-making, our experiments employed a quantified output from the guarding LLM. In this framework, the LLM produces a risk score between 0 and 10, allowing system administrators to specify the threshold at which messages are blocked. We conducted experiments using on-premise Llama models with $Llama3.1_{8B}$, $Llama3.3_{70B}$, and $Llama3.1_{405B}$ configurations. Each incoming message was evaluated ten times using the $Llama3.1_{8B}$ model, and three times with the other two models. This repeated evaluation was performed to capture worst-case scenarios, although a single evaluation would generally suffice in production environments. The maximum risk score observed for each message was recorded as the final outcome. For the $Llama3.1_{8B}$ model, as depicted in Fig. \ref{fig:fn-Llama31-8b}, the first false negative was observed at a decision threshold of 6. Threshold values of 9 and 10 were found to be insecure, as they allowed an excessive number of malicious messages to pass through.

    \begin{figure}
        \centering
        \includegraphics[scale=0.48]{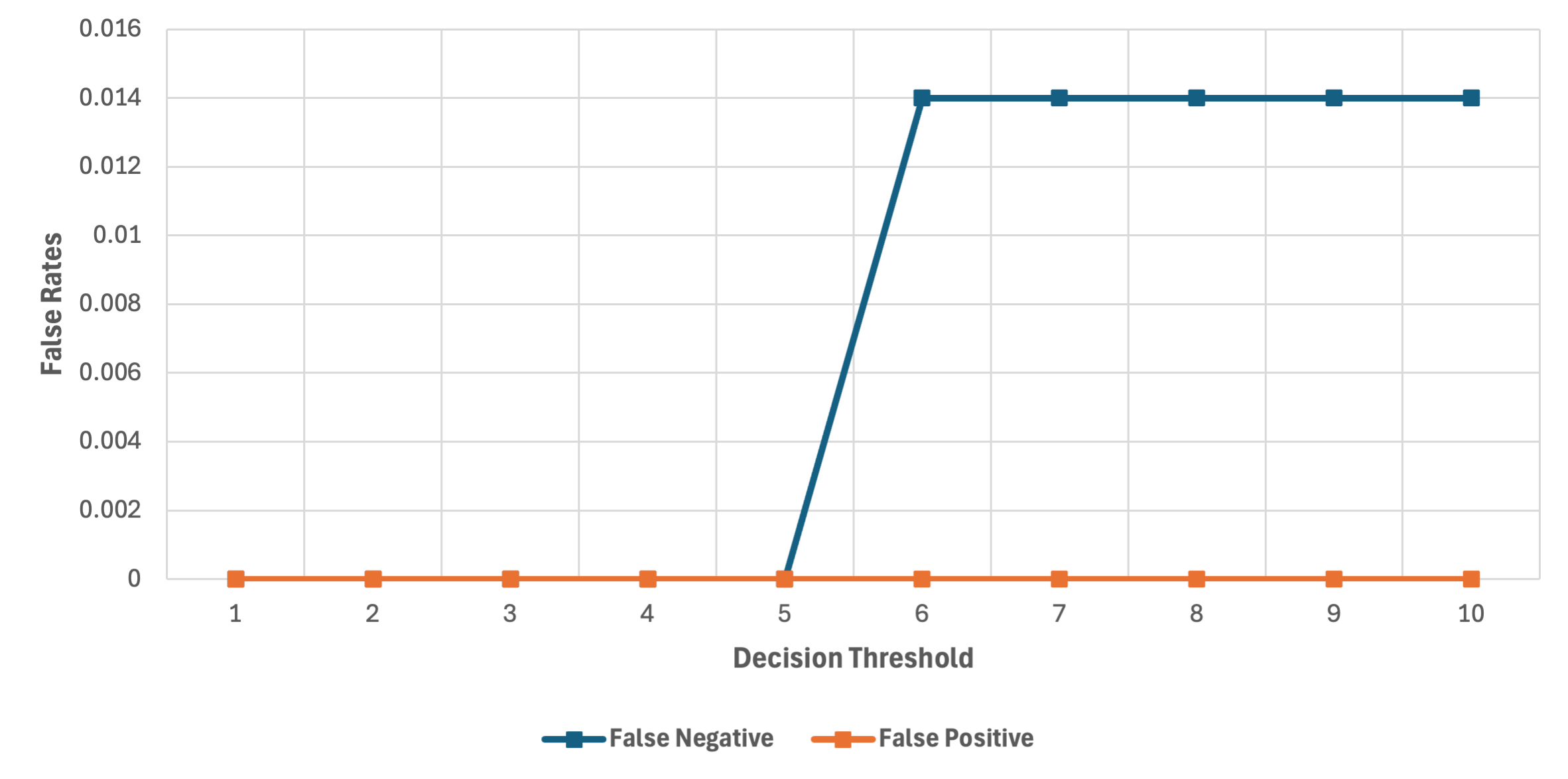}
        \caption{False rates, $Llama3.3_{70B}$, $Llama3.1_{405B}$, $C_{tot}$ = 71}
        \label{fig:fn-Llama33}
    \end{figure}

    As anticipated, transitioning from the $Llama3.1_{8B}$ model to the $Llama3.3_{70B}$ model resulted in a marked improvement in detection performance, effectively reducing the false positive rate to zero, as shown in Fig. \ref{fig:fn-Llama33}. This enhancement can be attributed to the design of our guard prompt, which requires sufficient model capacity to enforce rigorous message filtering. The evaluation process involves two distinct layers: the guard prompt and the subject message. The overall prompt length, along with the complexity of these layers, may exceed the processing capabilities of the $Llama3.1_{8B}$ model. Nonetheless, there is an optimal range of decision thresholds between 6 and 7 for $Llama3.1_{8B}$, which balances the detection of malicious content with the minimization of false positives. For the $Llama3.3_{70B}$ and $Llama3.1_{405B}$ models, a broader range of thresholds from 1 to 5 yields a false positive rate of zero. This flexibility enables fine-tuning of the pass-through criteria based primarily on the desired level of sensitivity for malicious message detection.

    In practical business deployments, LLMZ+ is utilized in combination with a simple message pre-processing step. As described in the following section, this configuration achieves ideal performance, resulting in both false positive and false negative rates of zero across all tested threshold values. Furthermore, we have successfully implemented LLMZ+ with various cloud-based models, including Google’s Gemini and its agentic framework.

\section{Practical Considerations}
    
    When deployed in a production environment, LLMZ+ must satisfy specific performance requirements to avoid disrupting the end-user experience. As demonstrated in the Results section, both the $Llama3.3_{70B}$ and $Llama3.1_{405B}$ models provide near-perfect detection performance. However, in resource-constrained settings, their execution times may be prohibitive. The following considerations outline methods to improve the efficiency of $Llama3.1_{8B}$ deployments.

    \subsection{False Positive Overrides}
    
    The majority of false positive cases observed with $Llama3.1_{8B}$ models stem from an incomplete understanding of the guard prompt. In particular, instructions related to associating risk ratings exclusively with prompt attack conditions, rather than with general data sensitivity, tend to be overlooked by models with fewer parameters. When evaluating LLMZ+ against real-world chatbot transcripts, we found that most false positives could be attributed to a limited set of frequently encountered message types, such as social security numbers, dates (including date of birth), and addresses. These messages can be detected and bypassed by a simple non-LLM filter, significantly reducing the false positive rate.

    \subsection{Message Pre-processing}

    Certain characteristics of incoming messages can be efficiently screened using traditional filters. One of the most effective checks is to impose a maximum message length. Prompt injection and related LLM exploit techniques typically require messages of considerable length to encode their malicious instructions. Such instructions often involve complex role redefinitions for the agentic model or the inclusion of encoding and decoding steps that enable the transfer of unauthorized content. By limiting the permissible message length, the system can block the vast majority of these attacks. In our experiments, a combination of message length filtering and use of the $Llama3.3_{70B}$ model resulted in both false positive and false negative rates of zero across all decision threshold values from 1 to 10. In this configuration, all malicious messages received a risk score of 10, while all legitimate messages were assigned a score of 0.

    \subsection{Parallel Execution}
    In applications where response time is critical, such as voice call interfaces, the system can be architected to execute the Guard prompt and the Agentic prompt simultaneously. This approach resembles branch prediction in CPUs, where future instructions are pre-processed in anticipation of conditional logic outcomes. In this configuration, the agentic response is withheld until the LLMZ+ decision is available. In the majority of cases, the LLMZ+ output is returned before the agentic response, resulting in improved overall system latency. However, this approach has certain drawbacks.

    \begin{itemize}
        \item It requires double the processing resources since both LLMs operate concurrently.
        \item In addition, there is a minor risk that an attacker could initiate a malicious agentic action prior to the completion of the LLMZ+ evaluation, although this is rare because most attacks focus on unauthorized data extraction rather than on immediate system disruption.
    \end{itemize}
    
    \subsection{Guard model selection}

    Although our findings demonstrate superior performance from the $Llama3.3_{70B}$ and $Llama3.1_{405B}$ models, model selection should be tailored to the specific deployment scenario. When LLMZ+ is not executed in parallel, it introduces a small synchronous delay to the agent’s response time. This delay may be exacerbated in agentic scenarios that require background activities, such as API calls, before delivering a response to the user. In some cases, it may be preferable to deploy a smaller model, such as $Llama3.1_{8B}$, and fine-tune it using the techniques described above. In real-time voice applications, for example, excessive response latency may cause users to terminate the interaction, which could negatively impact business operations. LLMZ+ is not intended as a one-size-fits-all solution, and careful model selection remains a key consideration for successful deployment.

\section{Conclusion and Future Work} 
    
    In this work, we introduced LLMZ+, a guard solution that filters both input to and output from agentic large language models. Conventional filtration methods generally focus on detecting malicious activity, which requires frequent updates to keyword lists or repeated model retraining as new prompt jailbreaking techniques emerge. In contrast, LLMZ+ is inspired by the $DENY ALL$ strategy commonly employed in firewall configurations and implements a dynamic whitelist approach that does not require retraining. The system specifically identifies compliant user prompts and agent responses, blocking all other content by default. Our approach is both straightforward and effective, particularly when deployed with larger LLMs, and achieved near-perfect detection rates with $Llama3.3_{70B}$ and $Llama3.1_{405B}$ models when evaluated against a set of the most recent jailbreak attacks.
    
    Future research can expand on this work by integrating a contextual Retrieval-Augmented Generation (RAG) pipeline to enhance LLMZ+'s assessment of agent responses. Additional efforts might focus on embedding content ring-fencing mechanisms directly into the LLM engine, further strengthening the protocol and making prompt injection attacks significantly more difficult to execute. Overall, LLMZ+ represents a meaningful advancement in the security of agentic AI systems.

\bibliographystyle{ieeetr}
\bibliography{references}

\end{document}